\def\year{2022}\relax
\documentclass[letterpaper]{article} 
\usepackage{aaai22}  
\usepackage{times}  
\usepackage{helvet}  
\usepackage{courier}  
\usepackage[hyphens]{url}  
\usepackage{graphicx} 
\usepackage{natbib}  
\usepackage{caption} 
\DeclareCaptionStyle{ruled}{labelfont=normalfont,labelsep=colon,strut=off} 
\usepackage{algorithm}
\usepackage{algorithmic}
\usepackage{multirow}

\usepackage{amsmath,amsfonts,bm}

\def\figref#1{figure~\ref{#1}}

\def\eqref#1{equation~\ref{#1}}

\def\1{\bm{1}}

\newcommand{\test}{\mathcal{D_{\mathrm{test}}}}

\DeclareMathAlphabet{\mathsfit}{\encodingdefault}{\sfdefault}{m}{sl}
\SetMathAlphabet{\mathsfit}{bold}{\encodingdefault}{\sfdefault}{bx}{n}

\newcommand{\N}{\mathcal{N}}
\newcommand{\E}{\mathbb{E}}
\newcommand{\V}{\mathbb{V}}

\DeclareMathOperator*{\argmax}{arg\,max}
\DeclareMathOperator*{\argmin}{arg\,min}

\newcommand{\thmref}[1]{Theorem~\ref{#1}}

\usepackage{xcolor}
\usepackage{paralist}
\usepackage{amsthm}

\usepackage{newfloat}
\usepackage{listings}
\floatstyle{ruled}
\newfloat{listing}{tb}{lst}{}
\floatname{listing}{Listing}
\renewcommand{\figref}[1]{Fig.~\ref{#1}}

\renewcommand{\thmref}[1]{Theorem~\ref{#1}}

\title{Training-Free Uncertainty Estimation for \\ Dense Regression: 
Sensitivity as a Surrogate}
\author{
Lu Mi\textsuperscript{\rm 1},
Hao Wang\textsuperscript{\rm 2},
Yonglong Tian\textsuperscript{\rm 1},
Hao He\textsuperscript{\rm 1},
Nir Shavit\textsuperscript{\rm 1}
}
\affiliations{
    \textsuperscript{\rm 1} MIT CSAIL\\
    \textsuperscript{\rm 2} Rutgers University\\
    lumi@mit.edu
}

\usepackage{bibentry}

\begin{document}

\maketitle

\def\Blue{\color{blue}}
\def\Purple{\color{purple}}

\def\A{{\bf A}}
\def\a{{\bf a}}
\def\B{{\bf B}}
\def\b{{\bf b}}
\def\C{{\bf C}}
\def\c{{\bf c}}
\def\D{{\bf D}}
\def\d{{\bf d}}
\def\e{{\bf e}}
\def\f{{\bf f}}
\def\F{{\bf F}}
\def\K{{\bf K}}
\def\k{{\bf k}}
\def\L{{\bf L}}
\def\H{{\bf H}}
\def\h{{\bf h}}
\def\G{{\bf G}}
\def\g{{\bf g}}
\def\I{{\bf I}}
\def\J{{\bf J}}
\def\X{{\bf X}}
\def\Y{{\bf Y}}
\def\OO{{\bf O}}
\def\oo{{\bf o}}
\def\P{{\bf P}}
\def\p{{\bf p}}
\def\Q{{\bf Q}}
\def\q{{\bf q}}
\def\r{{\bf r}}
\def\s{{\bf s}}
\def\S{{\bf S}}
\def\t{{\bf t}}
\def\T{{\bf T}}
\def\x{{\bf x}}
\def\y{{\bf y}}
\def\z{{\bf z}}
\def\Z{{\bf Z}}
\def\M{{\bf M}}
\def\m{{\bf m}}
\def\n{{\bf n}}
\def\U{{\bf U}}
\def\u{{\bf u}}
\def\v{{\bf v}}
\def\W{{\bf W}}
\def\w{{\bf w}}
\def\0{{\bf 0}}
\def\1{{\bf 1}}

\def\AM{{\mathcal A}}
\def\EM{{\mathcal E}}
\def\FM{{\mathcal F}}
\def\TM{{\mathcal T}}
\def\UM{{\mathcal U}}
\def\XM{{\mathcal X}}
\def\YM{{\mathcal Y}}
\def\NM{{\mathcal N}}
\def\OM{{\mathcal O}}
\def\IM{{\mathcal I}}
\def\GM{{\mathcal G}}
\def\PM{{\mathcal P}}
\def\LM{{\mathcal L}}
\def\MM{{\mathcal M}}
\def\DM{{\mathcal D}}
\def\SM{{\mathcal S}}
\def\ZM{{\mathcal Z}}
\def\RB{{\mathbb R}}
\def\EB{{\mathbb E}}
\def\VB{{\mathbb V}}

\def\tx{\tilde{\bf x}}
\def\ty{\tilde{\bf y}}
\def\tz{\tilde{\bf z}}
\def\hd{\hat{d}}
\def\HD{\hat{\bf D}}
\def\hx{\hat{\bf x}}
\def\hR{\hat{R}}

\def\Ome{\mbox{\boldmath$\omega$\unboldmath}}
\def\Om{\mbox{\boldmath$\Omega$\unboldmath}}
\def\bet{\mbox{\boldmath$\beta$\unboldmath}}
\def\et{\mbox{\boldmath$\eta$\unboldmath}}
\def\ep{\mbox{\boldmath$\epsilon$\unboldmath}}
\def\ph{\mbox{\boldmath$\phi$\unboldmath}}
\def\Pii{\mbox{\boldmath$\Pi$\unboldmath}}
\def\pii{\mbox{\boldmath$\pi$\unboldmath}}
\def\Ph{\mbox{\boldmath$\Phi$\unboldmath}}
\def\Ps{\mbox{\boldmath$\Psi$\unboldmath}}
\def\tha{\mbox{\boldmath$\theta$\unboldmath}}
\def\Tha{\mbox{\boldmath$\Theta$\unboldmath}}
\def\muu{\mbox{\boldmath$\mu$\unboldmath}}
\def\Si{\mbox{\boldmath$\Sigma$\unboldmath}}
\def\si{\mbox{\boldmath$\sigma$\unboldmath}}
\def\Gam{\mbox{\boldmath$\Gamma$\unboldmath}}
\def\gamm{\mbox{\boldmath$\gamma$\unboldmath}}
\def\Lam{\mbox{\boldmath$\Lambda$\unboldmath}}
\def\De{\mbox{\boldmath$\Delta$\unboldmath}}
\def\vps{\mbox{\boldmath$\varepsilon$\unboldmath}}
\def\Up{\mbox{\boldmath$\Upsilon$\unboldmath}}
\def\xii{\mbox{\boldmath$\xi$\unboldmath}}
\def\Xii{\mbox{\boldmath$\Xi$\unboldmath}}
\def\Lap{\mbox{\boldmath$\LM$\unboldmath}}
\newcommand{\ti}[1]{\tilde{#1}}

\def\tr{\mathrm{tr}}
\def\etr{\mathrm{etr}}
\def\etal{{\em et al.\/}\,}
\newcommand{\indep}{{\;\bot\!\!\!\!\!\!\bot\;}}
\def\argmax{\mathop{\rm argmax}}
\def\argmin{\mathop{\rm argmin}}
\def\vec{\text{vec}}
\def\cov{\text{cov}}
\def\dg{\text{diag}}

\newtheorem{observation}{\textbf{Observation}}
\newtheorem{remark}{Remark}
\newtheorem{theorem}{Theorem}
\newtheorem{lemma}{Lemma}
\newtheorem{definition}{Definition}
\newtheorem{problem}{Problem}
\newtheorem{proposition}{Proposition}
\newtheorem{cor}{Corollary}
\numberwithin{remark}{section}
\numberwithin{cor}{section}
\numberwithin{proposition}{section}

\begin{abstract}
Uncertainty estimation is an essential step in the evaluation of the robustness for deep learning models in computer vision, especially when applied in risk-sensitive areas. However, most state-of-the-art deep learning models either fail to obtain uncertainty estimation or need significant modification (e.g., formulating a proper Bayesian treatment) to obtain it. Most previous methods are not able to take an arbitrary model off the shelf and generate uncertainty estimation without retraining or redesigning it. \textcolor{black}{To address this gap, we perform a systematic exploration into training-free uncertainty estimation for dense regression, \textcolor{black}{an unrecognized yet important problem, and} provide a theoretical construction justifying such estimations.} 
We propose three simple and scalable methods to analyze the variance of outputs from a trained network under tolerable perturbations: \textit{infer-transformation}, \textit{infer-noise}, and \textit{infer-dropout}. They operate solely during the inference, without the need to re-train, re-design, or fine-tune the models, as typically required by state-of-the-art uncertainty estimation methods. Surprisingly, even without involving such perturbations in training, our methods produce comparable or even better uncertainty estimation when compared to training-required state-of-the-art methods. 

\end{abstract}
\section{Introduction}
\label{sec: intro}

Deep neural networks have achieved remarkable or even super-human performance in many tasks~\citep{krizhevsky2012imagenet,he2015delving,silver2016mastering}. While most previous work in the field has focused on improving accuracy in various tasks, in several risk-sensitive areas such as autonomous driving~\citep{chen2015deepdriving} and healthcare~\citep{zhang2019reducing}, reliability and robustness are arguably more important and interesting than accuracy. 

Recently, several novel approaches have been proposed to take into account an estimation of uncertainty during training and inference~\citep{huang2018efficient}. Some use probabilistic formulations for neural networks~\citep{graves2011,PBP,NPN,maxNPN} and model the distribution over the parameters (weights) and/or the neurons. Such formulations naturally produce distributions over the possible outputs~\citep{AUSE,yang2019inferring}. Others utilize the randomness induced during training and inference (e.g., dropout and ensembling) to obtain an uncertainty estimation~\citep{gal2016dropout,deepensemble,kendall2015bayesian}. 

All methods above require specific designs or a special training pipeline in order to involve the uncertainty estimation during training. Unfortunately, there are many cases where such premeditated designs or pipelines cannot be implemented. For example, if one wants to study the uncertainty of trained models released online, retraining is not always an option, especially when only a black-box model is provided or the training data is not available. Moreover, most models are deterministic and do not have stochasticity. A straightforward solution is to add dropout layers into proper locations and finetune the model~\citep{gal2016dropout}. However, this is impractical for many state-of-the-art and published models, especially those trained on large datasets (e.g. ImageNet~\citep{deng2009imagenet}) with a vast amount of industrial computing resources. In addition, models that have already been distilled, pruned, or binarized fall short of fitting re-training~\citep{deepcompression,binaryNN}. 

\textcolor{black}{To fill this gap, we identify the problem of \emph{training-free uncertainty estimation}: how to obtain an uncertainty estimation of any given model without re-designing, re-training, or fine-tuning it.} We focus on two scenarios: black-box uncertainty estimation (BBUE), where one has access to the model only as a black box, and gray-box uncertainty estimation (GBUE), where one has access to intermediate-layer neurons of the model (but not the parameters). \textcolor{black}{
Our work is a systematic exploration of this unrecognized yet important problem.}

We propose a set of simple and scalable training-free methods to analyze the variance of the output from a trained network, shown in \figref{fig:method_description}. Our main idea is to add a \textit{tolerable} perturbation into inputs or feature maps during inference and use the variance of the output as a surrogate for uncertainty estimation. 

The first method, which we call \emph{infer-transformation}, is to apply a transformation that exploits the natural characteristics of a CNN -- it is variant to input transformation such as rotation~\citep{cohen2016group}. Transformations have been frequently used as data augmentation but rarely evaluated for uncertainty estimation. The second method, \emph{infer-noise}, is to inject Gaussian noise with a zero-mean and a small standard deviation into intermediate-layer neurons. The third one, called \emph{infer-dropout}, is to perform inference-time dropout in a chosen layer. Although at first blush infer-dropout is similar to MC-dropout, where dropout is performed during both training and inference in the same layers, they are different in several aspects: 
(1) Infer-dropout is involved \emph{only during inference}. (2) Infer-dropout can be applied to arbitrary layers, even those without dropout training. Surprisingly, we find that even without involving dropout during training, infer-dropout is still comparable to, or even better than, MC-dropout for the purpose of uncertainty estimation.

In our paper, we focus on regression tasks. Note that for classification tasks, the softmax output is naturally a distribution. Methods that use entropy for uncertainty estimation qualify as a training-free method and have outperformed MC-Dropout~\citep{bahat2018confidence,gal2016dropout,hendrycks2016baseline,wang2019aleatoric} (see the Supplement
for experiment results). Regression tasks are more challenging than classification problems since there is no direct output distribution.~\citep{kuleshov2018accurate, song2019distribution}. And our major contributions are: 

\begin{compactenum}

\item We perform a systematic exploration of training-free uncertainty estimation for regression models \textcolor{black}{and provide a theoretical construction justifying such estimations}.

\item \textcolor{black}{We propose simple and scalable methods, \emph{infer-transformation}, \emph{infer-noise} and \emph{infer-dropout}, using a tolerable perturbation to effectively and efficiently estimate uncertainty.}

\item Surprisingly, we find that our methods are able to generate uncertainty estimation \textcolor{black}{comparable or even better than~\textcolor{black}{training-required} baselines}
in~\textcolor{black}{real-world} large-scale \textcolor{black}{dense} regression tasks.

\end{compactenum}

\section{Related Work}

\label{sec: related}

\textbf{Probabilistic Neural Networks for Uncertainty Estimation.} Probabilistic neural networks consider the input and model parameters as random variables which take effect as the source of stochasticity \cite{nix1994estimating,welling2011bayesian,graves2011,PBP,NPN}. Traditional Bayesian neural networks model the distribution over the parameters (weights) \cite{mackay1992,hinton1993,graves2011,welling2011bayesian} and obtain the output distribution by marginalizing out the parameters. Even with recent improvement \cite{BDK,PBP}, one major limitation is that the size of network at least doubles under this assumption, and the propagation with a distribution is usually computationally expensive. Another set of popular and efficient methods~\cite{gal2016dropout,teye2018bayesian} formulate dropout~\cite{srivastava2014dropout} or batch normalization~\cite{ioffe2015batch} as approximations to Bayesian neural networks. For example, MC-dropout~\cite{gal2016dropout} injects \textcolor{black}{dropout} 
into some layers during both training and inference ~\cite{tsymbal}. Unlike most models that disable dropout during inference, MC-dropout feed-forwards the same example multiple times with dropout enabled, in order to form a distribution on the output. Meanwhile, other works~\cite{NPN,maxNPN} propose sampling-free probabilistic neural networks as a lightweight Bayesian treatment for neural networks.

\textbf{Non-probabilistic Neural Networks for Uncertainty Estimation.} Other strategies~\cite{zhao2020individual} such as deep ensemble~\cite{deepensemble,huang2017snapshot,ashukha2020pitfalls} train an ensemble of neural networks from scratch, where some randomness is induced during the training process, i.e. the initial weight is randomly sampled from a distribution. During inference, these networks will generate a distribution of the output. Though simple and effective, training multiple networks costs even more time and memory than Bayesian neural networks. 
Another efficient method log likelihood maximization (LLM) is to train the network to have both original outputs and uncertainty predictions, by jointly optimizing both~\cite{zhang2019reducing, poggi2020uncertainty}. Besides the methods above focusing on uncertainty in classification models; there are also works investigating uncertainty in regression models~\cite{kuleshov2018accurate,song2019distribution,zelikman2020crude}. 
However, all methods above requires re-training, introduces heavy implementation overhead, and sometimes makes the optimization process more challenging. 


\section{Methodology}
\label{sec: method}

\begin{figure*}[t]
\begin{center}

\includegraphics[width=0.75\linewidth]{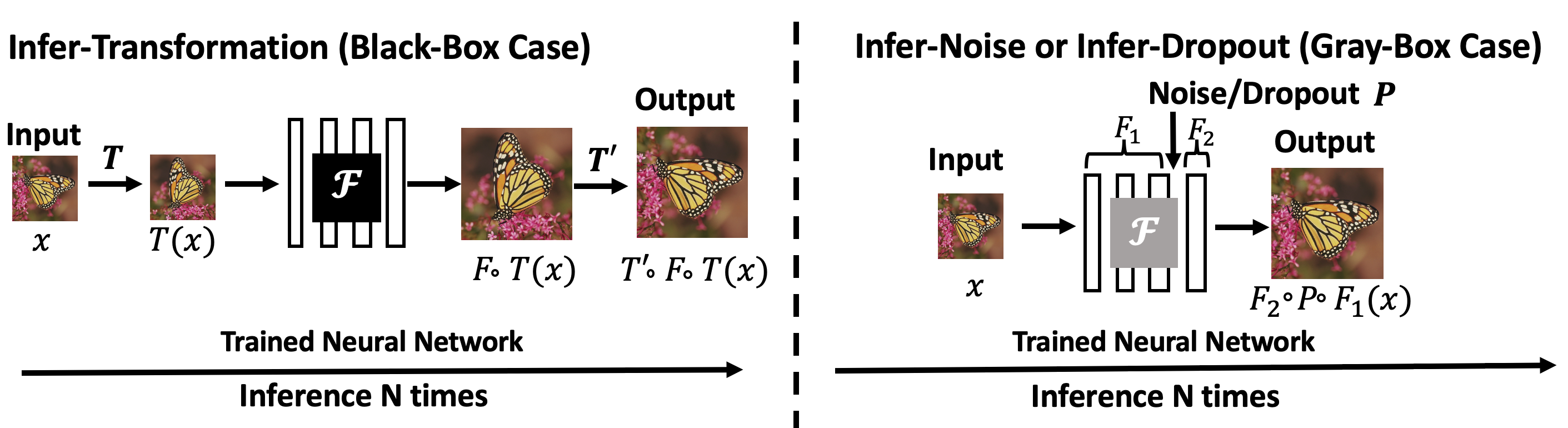}
\end{center}
\vspace{-10pt}
\caption{Method description of our training-free uncertainty estimation: apply infer-transformation $T$ (\textbf{left}) and infer-noise or infer-dropout $P$ (\textbf{right}) to a trained neural network $F$ during inference.}
\vspace{-10pt}
\label{fig:method_description}

\end{figure*}

\textbf{Three Cases on Parameter Accessibility.}
We distinguish among three cases based on accessibility of the original model. 1. Black-box case: the model is given as a trained black box without any access to its internal structure. 2. Gray-box case: the internal representations (feature maps) of the model is accessible (while the parameters are not) and can be modified during inference. 3. White-box case: the model is available for all modifications (e.g. its weights can be modified,  \textcolor{black}{which requires training}). In this paper we focus on the black-box and gray-box cases, for which we offer, correspondingly, two classes of methods. 
For the black-box case, we propose \emph{infer-transformation}, which exploits the model's dependence on input transformations, e.g. rotations/flips. For the grey-box case, we propose \emph{infer-noise} and \emph{infer-dropout}, which introduce an internal embedding/representation manipulation - injecting a noise layer or a dropout layer during inference. These three methods are illustrated in \figref{fig:method_description}. The description of our methods and a \textcolor{black}{theoretical construction} are presented as below.

\subsection{Black-Box Uncertainty Estimation: Infer-Transformation}

Given a black-box model, we explore the behavior of the outputs for different transformed versions of the input. Specifically, we transform the input with tolerable perturbations, e.g. perturbations that do not cause significant increase in the loss, and then use the variance of the perturbed outputs as estimated uncertainty. Here we focus on transformations that preserve pertinent characteristics of the input, such as rotations, flips, etc. Formally, given an input image X, our measured uncertainty is defined as $\mathbb{V}[Z] = \mathbb{V}_T[T'\circ F \circ T(X)]$, where $T\in \mathcal{T}$ is a transformation, $T'$ is $T$'s inverse operation, and $F$ is a function representing the black-box neural network. $Z=T'\circ F \circ T(X)$ is a sample from the perturbed output distribution. Note that it is possible to sample $Z=F(X)$, where $T$ happens to be a 360-degree rotation.

\begin{figure*}[ht]
\vspace{-7pt}
\centering

	 \includegraphics[width=0.8\linewidth]{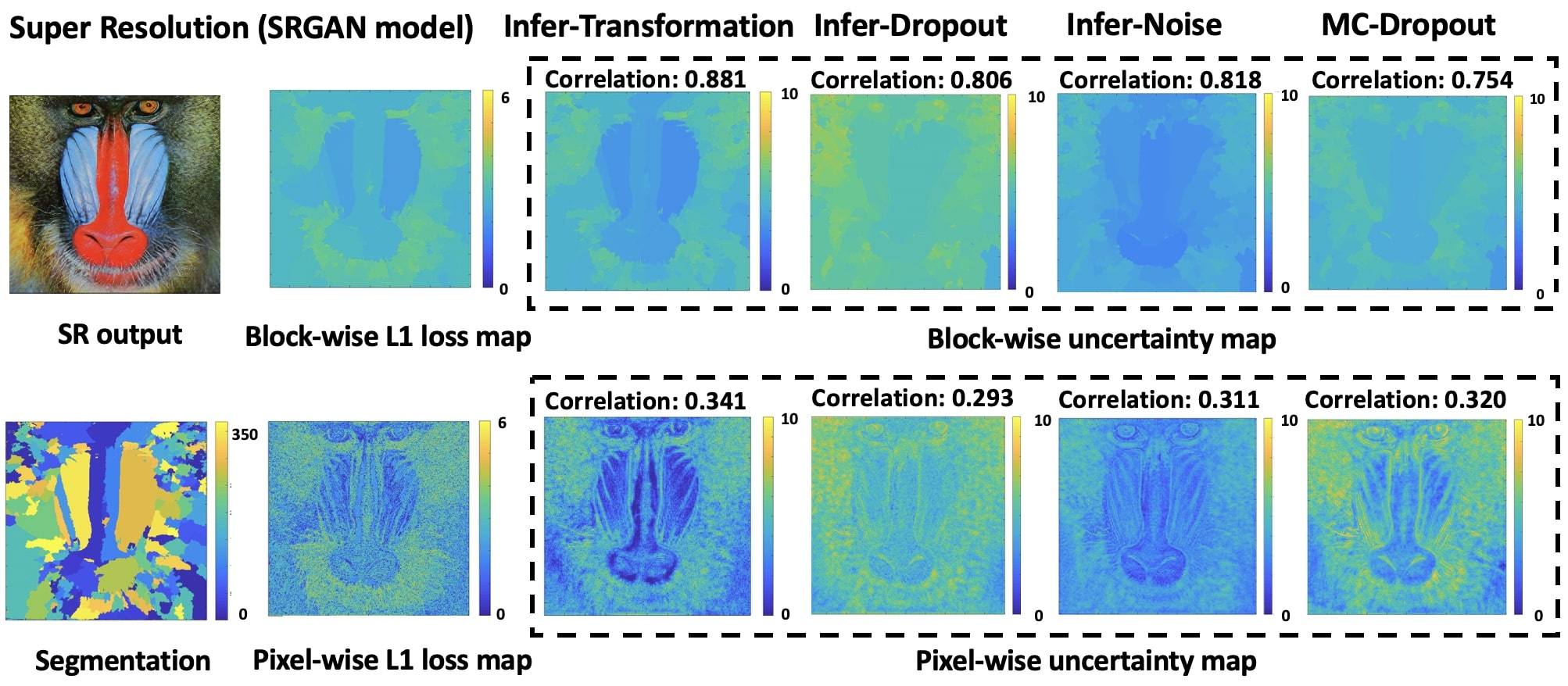}

    \caption{Visualization of block-wise and pixel-wise uncertainty (variance) maps (log scale) generated by infer-transformation, infer-dropout, MC-dropout~\citep{gal2016dropout}, using SRGAN~\citep{ledig2017photo} for the super resolution task.
    $L_1$ loss map (log scale) is also provided for comparison. Correlation between \textcolor{black}{the $L_1$ loss map} and the uncertainty map is presented. }

\label{fig:variance_map}
\end{figure*}

\subsection{Gray-Box Uncertainty Estimation: Infer-Noise and Infer-Dropout}
Given a gray-box model, we consider another class of methods for generating multiple outputs from a distribution: randomly perturbing latent codes. \textcolor{black}{Compared with the black-box case, this provides finer granularity on modulating the perturbation strength to ensure tolerability.} 
Specifically, we propose \emph{infer-noise}, which introduces Gaussian noise at an intermediate layer of the trained model, and \emph{infer-dropout}, which uses dropout instead. For infer-noise, the noise will be added to the feature maps of a certain layer. 
This noise is randomly sampled multiple times during inference to form a set of diverse outputs. 
For infer-dropout, 
random dropout is performed for multiple forwards to generate output samples, the variance of which is then used as uncertainty estimation. Formally, given an input image X, our measured uncertainty is defined as $\mathbb{V}[Z] = \mathbb{V}_P[F_2 \circ P\circ F_1(X)]$, \textcolor{black}{where $P$ is sampled from a perturbation set $\mathcal{P}$ (e.g. Gaussian noise with $\sigma=1$). $F_1$ is the function of network layers before the perturbation $P$, $F_2$ represents network layers after $P$, and $F_2\circ F_1(X)$ is the gray-box network $F(X)$.} 
$Z=F_2 \circ P\circ F_1(X)$ is a sample from the perturbed output distribution. Note that it is possible to sample $Z=F(X)$, where $P$ happens to be a perturbation noise of all zeros.
{
\subsection{Correlation between Sensitivity and Uncertainty}
}
{
In this section, we provide a theoretical justification for using sensitivity as a surrogate of uncertainty. Specifically, we will show that sensitivity and uncertainty have a non-negative correlation in a simplified setting with mild assumptions. }

{
\textbf{Notations.}
We use random variables $X$ and $Y$ to denote the input data point and target label. $Z_0=F_2\circ F_1(X)$ and $Z=F_2\circ P \circ F_1(X)$ denote the model predictions of original and perturbed input, respectively. Furthermore, $\mu(X)=\E_{P}[Z|X]$ and $\sigma(X)^2=\V_{P}[Z|X]$ denote the expectation and the variance of the model prediction of perturbations of a certain input $X$. We call $\sigma(X)$ (or $\sigma$ for short) the sensitivity of the model for the input $X$. We use $e=|Z_0-Y|$ to denote the error of the model's prediction. We consider $e$ as an indicator of the model's uncertainty: a larger prediction error means that the model should have been less uncertain at this input. We use $\rho(\cdot,\cdot)$ to denote the Pearson correlation between two random variables. 
}
\newcommand{\eone}{\epsilon_1}
\newcommand{\etwo}{\epsilon_2}

\textbf{Assumptions.} To analyze the correlation between sensitivity and uncertainty, we make the following assumptions:
\begin{enumerate}
\item Heterogeneous perturbation: $\epsilon_1 = \frac{Z_0 - \mu(X)}{\sigma(X)} \sim \N(0,1)$. This assumption says the model prediction given an unperturbed input `looks like' a random draw from the model predictions of perturbed inputs.
\item Random bias: $\epsilon_2 = Y - \mu(X) \sim \N(0,B^2)$. This assumption says the bias of the model prediction `looks like' white noise with bounded variance $B^2$.
\item Independence: $\eone \indep \etwo$ and $(\eone,\etwo) \indep \sigma$. Basically, we assume the randomness in $\eone$, $\etwo$ and the sensitivity $\sigma$ are statistically independent. 
\end{enumerate}
{
Empirically, we find these assumptions are satisfied to some degree (see detailed results in the Supplement).  
}

\textbf{Correlation Analysis.} The following theorem analyzes the correlation between sensitivity $\sigma$ and uncertainty $e$.
\begin{theorem}\label{thm:corr}
With the above assumptions satisfied, we have:
\begin{equation}
\rho(\sigma,e) = \sqrt{\frac{2}{\pi} \frac{(1-\lambda^2)}{(1 - \frac{2}{\pi} \lambda^2)}} \rho(\sigma, \sigma_B)
\end{equation}where $\sigma_B=\sqrt{\sigma^2+B^2}$ and $\lambda^2=\frac{(\E [\sigma_B])^2}{\E [\sigma_B^2]}$.
\end{theorem}
We make several remarks on the derived correlation between sensitivity and uncertainty. 
\begin{enumerate}
\item $\rho(\sigma,e) \geq 0$: It holds because $\sigma$ and $\sigma_B$ always have non-negative correlation. It means statistically, sensitivity has the same `direction' as uncertainty. 
\item $\rho(\sigma,e)$ is monotonically decreasing w.r.t. to $B$: It holds since $\rho(\sigma,\sigma_B)$ decreases as $B$ grows. Consider two extremes: (a) $B=0$ (the prediction has no bias at all): $\sigma_B$ then degenerates to $\sigma$, leading to the highest sensitivity-uncertainty correlation. 
(b) $B=\infty$ (the prediction has unbounded bias). $\sigma_B$ then degenerates to $B$, leading to zero sensitivity-uncertainty correlation. 
\item $\rho(\sigma,e)$ is monotonically decreasing w.r.t. to $\lambda^2$: It holds since $\frac{1-\lambda^2}{1-\frac{2}{\pi}\lambda^2}$ decreases as $\lambda^2$ grows. Note that $\lambda^2$ is bounded (between $0$ and $1$) and that $\lambda^2$ indicates the variability of the sensitivity. Specifically, $\lambda^2=1$ is equivalent to $\V[\sigma]=0$, meaning that the sensitivity is constant everywhere. In this extreme case, sensitivity has zero correlation with uncertainty. Fortunately, in practice, we find sensitivity always varies with different inputs. The variation is often large, leading $\lambda^2$ relatively small. As a result, sensitivity usually has a high correlation with uncertainty. 
\item $\rho(\sigma,e) \leq \sqrt{\frac{2}{\pi}} \approx 0.8$: Using sensitivity as a surrogate can at most achieve pixel-wise correlation of $\sqrt{2/\pi}$. For block-wise, patch-wise, and image-level correlation (see the Experiment Section for definitions), we can apply similar analysis to obtain higher upper bounds. 
\end{enumerate}
\thmref{thm:corr} establishes the correlation between sensitivity $\sigma$ and uncertainty $e$. In our experiments, we use $\sigma^2$ as sensitivity for convenience since it achieves similar performance. 

\subsection{Epistemic Uncertainty and Aleatoric Uncertainty} Our model can estimate both epistemic uncertainty and aleatoric Uncertainty~\citep{kendall2017uncertainties}. As in practice, the model is not able to perfectly fit infinite data -- all variants of augmented “data” (including both data inputs and intermediate features) applied with different perturbations -- then we will get a nonzero variance, which represents the epistemic uncertainty. In the meanwhile, our methods can also be applied to measure aleatoric uncertainty for accessible data. Given a data input fed into a trained model under multiple tolerable perturbations, outputs with higher variance than those from other data inputs represent this data input with relatively high aleatoric uncertainty. And we demonstrate to use of such properties to implement active learning, and results are shown in the Supplement.

\section{Experiments}\label{sec:exp}
\label{sec: exp}

\begin{figure*}[t]
\centering
\vspace{-7 pt}
\includegraphics[width=0.7\linewidth]{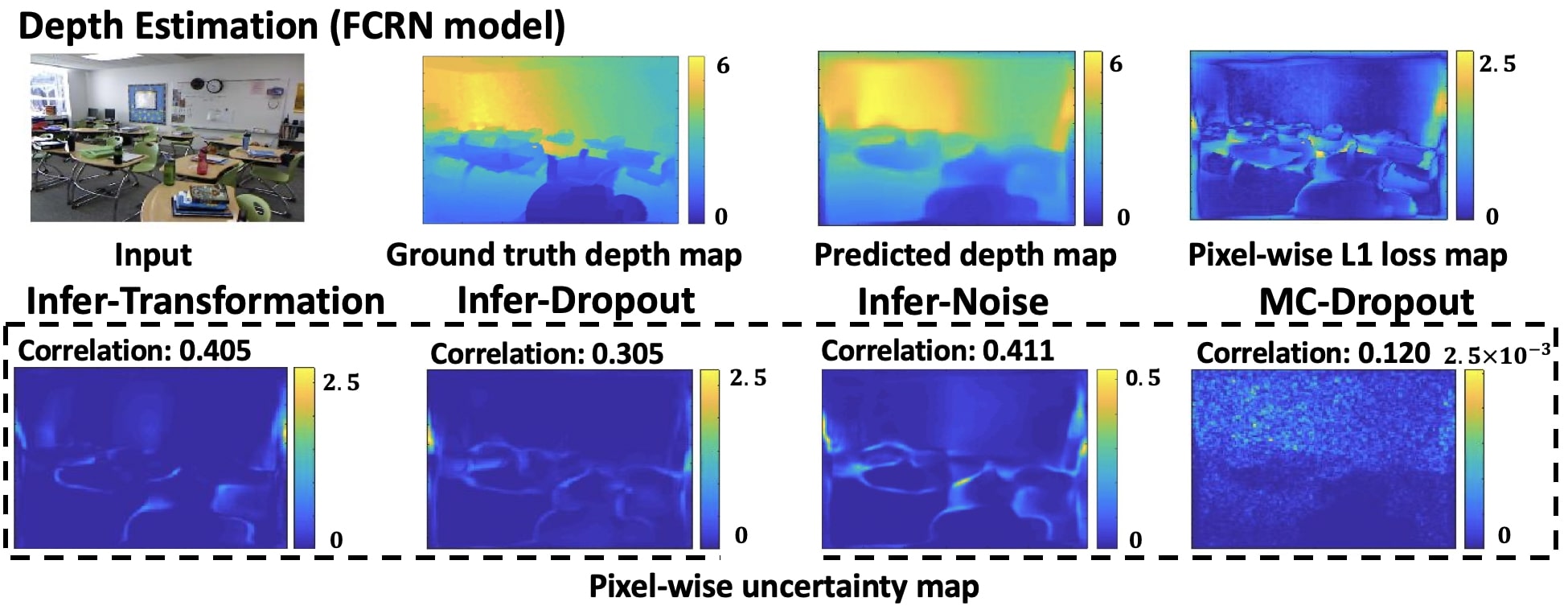}
\caption{Visualization of pixel-wise uncertainty (variance) maps from infer-transformation, infer-dropout, MC-dropout~\citep{gal2016dropout} compared with the $L_1$ loss map in depth estimation task. Correlation between the $L_1$ loss map and the uncertainty map is also presented.}
\vspace{-5pt}
\label{fig:variance_map_depth}
\end{figure*}

In this section, we evaluate our three proposed approaches in two \textcolor{black}{representative real-world large-scale dense regression} tasks, super single image resolution, and monocular depth estimation.
\subsection{Single Image Super Resolution}

The task of Single Image Super Resolution (SR) is to reconstruct a
high-resolution (HR) image from a low-resolution (LR) input. 
Here we focus on analyzing the state-of-the-art SRGAN model~\citep{ledig2017photo}, which can restore photo-realistic high-quality images. SRGAN always outputs deterministic restorations since the conditional GAN~\citep{mirza2014conditional} used in this model involves no latent variable sampling. However, we can still evaluate its uncertainty with our proposed methods.

We apply our methods to estimate uncertainty in one open-source version of this work~\citep{tensorlayer2017}. The package provides two models trained with different loss functions: 1) SRresnet model with $L_2$ loss and 2) SRGAN model with a combination of $L_2$ loss and adversarial loss. We evaluate our methods on both models in the black-box/gray-box settings.

\textbf{Infer-Transformation.} For infer-transformation, we apply a rotation of $K\times 90$ degrees ($K=0, 1,2,3$) as well as horizontal flip to the LR input, feed it into the trained model during the inference, and apply the inverse transformation to its output. \textcolor{black}{We could generate at most 8 samples using this strategy, and then calculate the pixel-wise variance.}

\textbf{Infer-Noise.} In infer-noise, we take the trained model and add a Gaussian-noise layer, which has standard deviation $\sigma \in \{0.01, 0.02, 0.05, 0.1, 0.2, 0.5\}$ and mean $0$, at different locations (layers). We choose 4 different locations for noise injection, including the layers right after the input and some intermediate layers \textcolor{black}{(see details in the Supplement}. 
For each experiment, we only add the noise into one layer with a specific $\sigma$ value. \textcolor{black}{Sample numbers of 8 and 32 are evaluated.}

\textbf{Infer-Dropout.} In infer-dropout, we take the trained model and add a dropout layer with varied dropout rates. We choose the dropout rate $\rho$ from the set $\{0.01, 0.02, 0.05, 0.1, 0.2, 0.5\}$ and use the same set of locations as the infer-noise. For each experiment, we only add the layer into one location with one specific dropout rate. \textcolor{black}{Sample numbers of 8 and 32 are evaluated.} 

\textbf{Baselines.} \textcolor{black}{We compare our methods with three training-required baselines}. The first baseline is MC-dropout~\citep{gal2016dropout} with a dropout rate $\rho \in \{0.01, 0.02, 0.05, 0.1, 0.2, 0.5\}$. For each experiment, we add a dropout layer only into one location with one dropout rate during training. The same dropout rate is used for sampling during inference. \textcolor{black}{We try different sample numbers of 8 and 32.} 
The second baseline is deep ensemble~\citep{deepensemble}. We follow this work to train ensembles as $4$ and $8$ networks, respectively. We train these networks with the same number of epochs until they converge. \textcolor{black}{During inference, each of them generates a single deterministic output, with $4$ or $8$ samples} generated in total. \textcolor{black}{The third baseline is a sampling-free method log-likelihood maximization (LLM) ~\citep{zhang2019reducing, poggi2020uncertainty}, where a network is trained to predict an output distribution with log-likelihood maximization}.

\subsection{Monocular Depth Estimation}

For depth estimation~\citep{postels2019sampling,kendall2017uncertainties}, we use one of the commonly applied models based on a fully convolutional residual network (FCRN)~\citep{laina2016deeper}.
We directly use the trained model released by the original author; this is consistent with the scenarios of black-box and gray-box cases since the code for training is not released. We evaluate the model on NYU Depth Dataset V2. \textcolor{black}{For infer-transformation, we avoid applying 90-degree rotation to input, \textcolor{black}{since the orientation is strong prior to predicting depth} which can violate the tolerability, and only apply horizontal flip to generate $2$ samples for uncertainty estimation.} For infer-dropout, we choose two locations (intermediate layers) to add the dropout layer. For infer-noise, we choose three locations to add the noise layer (two intermediate layers and one layer before the final FC layer). Then we conduct similar experiments as described in the SR task. For the baseline MC-dropout, note that the model has a dropout layer before the final fully connected (FC) layer during training, we directly perform sampling from the existing dropout layer. Sample numbers of $2$ and $8$ are evaluated for both infer-dropout and infer-noise, see details in \textcolor{black}{the Supplement}. 

\begin{table*}[t]
\setlength{\tabcolsep}{5pt}
\begin{center}
\footnotesize
\vspace{-6pt}
\begin{tabular}{c|c|cccccc|ccccccc}
\hline
\multicolumn{15}{c}{SRGAN model: Super Resolution}\\
\hline
\multicolumn{2}{c|}{Condition} &\multicolumn{6}{c}{Training Free (Ours)} &\multicolumn{6}{|c}{Training Required}\\
\hline
\multicolumn{2}{c|}{Prediction} &\multicolumn{8}{c}{Original Output} &\multicolumn{5}{|c}{Outputs Mean}\\ 
\hline
\multicolumn{2}{c}{Method} &\multicolumn{2}{|c}{Infer-trans} &  \multicolumn{2}{|c}{Infer-drop} & \multicolumn{2}{|c}{Infer-noise} & \multicolumn{2}{|c}{MC-drop$^1$}& \multicolumn{2}{|c|}{MC-drop$^2$} &\multicolumn{2}{|c}{Ensemble}
&\multicolumn{1}{|c}{LLM}
\\
\hline
\multicolumn{2}{c|}{Samples} & 4 &8 &8 &32 &8 &32 &8 &32 &8 &32 &4 &8 &0\\
\hline
\multirow{4}{*}{AUSE} &mean $L_1$ &\textbf{0.006}	&\textbf{0.006}	&0.013	&0.013	&0.016	&0.016	&0.008	&0.009	&0.007	&0.007	&0.018	&0.020 &0.035\\

&patch $L_1$ &\textbf{0.021}	&0.022	&0.029	&0.030	&0.040	&0.039	&0.029	&0.029	&0.025	&0.025	&0.033	&0.033 &0.044\\

&block $L_1$ &\textbf{0.023}	&\textbf{0.023}	&0.032	&0.031	&0.044	&0.043	&0.030	&0.029	&0.028	&0.028	&0.033	&0.033 &0.042\\

&pixel $L_1$ &0.128	&\textbf{0.121}	&0.154	&0.141	&0.173	&0.162	&0.141	&0.131	&0.137	&0.129	&0.152	&0.144 &0.137\\
\hline
\multirow{4}{*}{Corr} &mean $L_1$ &0.931	&0.930	&0.884	&0.882	&0.774	&0.780	&0.942	&\textbf{0.943}	&0.938	&0.936	&0.692	&0.694 &0.484\\

&patch $L_1$ &0.765	&\textbf{0.770}	&0.722	&0.731	&0.590	&0.598	&0.748	&0.755	&0.734	&0.741	&0.674	&0.677 &0.565\\

&block $L_1$ &0.757	&\textbf{0.767}	&0.717	&0.730	&0.579	&0.592	&0.735	&0.747	&0.698	&0.710	&0.651	&0.664 &0.588\\

&pixel $L_1$ &0.367	&\textbf{0.394}	&0.323	&0.376	&0.245	&0.288	&0.339	&0.390	&0.330	&0.379	&0.290	&0.326 &0.393\\
\hline
\multicolumn{2}{c|}{NLL} &17.910	&9.332	&4.889	&4.804	&4.899	&4.791	&6.804	&6.013	&6.365	&5.541	&11.520	&5.994 &\textbf{1.320}\\
\hline
\hline
\multicolumn{15}{c}{SRresnet model: 
Super Resolution}\\

\hline
\multicolumn{2}{c|}{Condition} &\multicolumn{6}{c}{Training Free (Ours)} &\multicolumn{6}{|c}{Training Required}\\
\hline
\multicolumn{2}{c|}{Prediction} &\multicolumn{8}{c}{Original Output} &\multicolumn{5}{|c}{Outputs Mean}\\ 

\hline
\multicolumn{2}{c}{Method} &\multicolumn{2}{|c}{Infer-trans} &  \multicolumn{2}{|c}{Infer-drop} & \multicolumn{2}{|c}{Infer-noise} & \multicolumn{2}{|c}{MC-drop$^1$}& \multicolumn{2}{|c}{MC-drop$^2$} &\multicolumn{2}{|c}{Ensemble}
&\multicolumn{1}{|c}{LLM}
\\
\hline
\multicolumn{2}{c|}{Samples} & 4 &8 &8 &32 &8 &32 &8 &32 &8 &32 &4 &8 &0\\
\hline
\multirow{4}{*}{AUSE} &mean $L_1$ &0.044	&0.044	&0.042	&0.043	&0.067	&0.066	&0.041	&0.047	&\textbf{0.036}	&0.039	&0.073	&0.066 &\textbf{0.036}\\

&patch $L_1$ &0.045	&0.044	&0.048	&0.047	&0.055	&0.055	&0.046	&0.045	&0.040	&0.041	&0.066	&0.066 &\textbf{0.037}\\

&block $L_1$ &0.047	&0.045	&0.049	&0.048	&0.062	&0.061	&0.048	&0.046	&0.041	&0.042	&0.073	&0.070 &\textbf{0.031}\\

&pixel $L_1$ &0.164	&0.153	&0.165	&0.155	&0.190	&0.181	&0.163	&0.152	&0.150	&0.144	&0.194	&0.185 &\textbf{0.133}\\
\hline
\multirow{4}{*}{Corr} &mean $L_1$ &0.340	&0.359	&0.401	&0.404	&0.056	&0.055	&0.408	&0.379	&0.527	&0.512	&0.016	&0.048 &\textbf{0.622}\\

&patch $L_1$ &0.501	&0.520	&0.508	&0.518	&0.371	&0.385	&0.535	&0.545	&0.547	&0.542	&0.323	&0.361 &\textbf{0.648}\\

&block $L_1$ &0.462	&0.486	&0.498	&0.509	&0.358	&0.370	&0.505	&0.521	&0.531	&0.529	&0.274	&0.286 &\textbf{0.673}\\

&pixel $L_1$ &0.237	&0.269	&0.258	&0.303	&0.172	&0.216	&0.264	&0.309	&0.288	&0.322	&0.184	&0.206 &\textbf{0.393}\\
\hline
\multicolumn{2}{c|}{NLL} &107.243	&43.071	&5.155	&4.955	&4.941	&4.788	&8.430	&7.221	&8.018	&6.688	&13.559	&7.906 &\textbf{1.422}\\
\hline
\end{tabular}
\end{center}
\caption{Mean/patch-wise/block-wise/pixel-wise AUSE and correlation between $L_1$ loss and uncertainty, and NLL on SR benchmark dataset Set 14. Our infer-transformation, infer-dropout and infer-noise are compared with MC-dropout~\citep{gal2016dropout}, deep ensemble~\citep{deepensemble}, and \textcolor{black}{log likelihood maximization (LLM) ~\citep{zhang2019reducing}}. MC-drop$^1$ uses the output of the original model as a prediction while MC-drop$^2$ uses the mean of output samples from the re-trained model (with added dropout) as prediction. Models evaluated: SRGAN and SRresnet.}
\label{Tab:correlation with error}
\end{table*}

\begin{table*}[h]
\setlength{\tabcolsep}{6pt}
\begin{center}
\footnotesize
\begin{tabular}{c|c|ccccc|cccc}
\hline
\multicolumn{11}{c}{FCRN model: Depth Estimation}\\
\hline
\multicolumn{2}{c|}{Condition} &\multicolumn{5}{c}{Training Free (Ours)} &\multicolumn{4}{|c}{Training Required}\\

\hline
\multicolumn{2}{c|}{Prediction} &\multicolumn{7}{c}{Original Output} &\multicolumn{2}{|c}{Outputs Mean}\\ 
\hline
\multicolumn{2}{c|}{Method} &\multicolumn{1}{c}{Infer-trans} &  \multicolumn{2}{|c}{Infer-drop} & \multicolumn{2}{|c}{Infer-noise} & \multicolumn{2}{|c}{MC-drop$^1$}& \multicolumn{2}{|c}{MC-drop$^2$}\\
\hline
\multicolumn{2}{c|}{Samples} &2 & 2 &8 &2 &8 &2 &8 &2 &8\\
\hline
\multirow{4}{*}{AUSE} &mean $L_1$ &0.051	&0.044	&\textbf{0.041}	&0.046	&\textbf{
0.041}	&0.062 &0.062 &0.060	&0.062\\

&patch $L_1$ &0.106	&0.109	&0.092	&0.108	&\textbf{0.091}	&0.127	&0.126	&0.122	&0.125\\

&block $L_1$ &0.056	&0.057	&0.047	&0.053	&\textbf{0.045}	&0.065	&0.065	&0.063	&0.065\\

&pixel $L_1$ &0.165	&0.168	&0.135	&0.167	&\textbf{0.134}	&0.208	&0.193	&0.207	&0.193\\
\hline
\multirow{4}{*}{Corr} &mean $L_1$ &0.596	&0.630	&0.677	&0.651	&\textbf{0.708}	&0.473	&0.471 &0.469	&0.469\\

&patch $L_1$ &0.324	&0.306	&0.409	&0.312	&\textbf{0.411}	&0.258	&0.268	&0.266	&0.269\\

&block $L_1$ &0.354	&0.354	&\textbf{0.449}	&0.364	&0.447	&0.215	&0.220	&0.220	&0.221\\

&pixel $L_1$ &0.208	&0.182	&0.284	&0.188	&\textbf{0.288}	&0.075	&0.134	&0.076	&0.134\\
\hline
\multicolumn{2}{c|}{NLL} &8.634	&8.443	&4.889	&3.526	&\textbf{1.006}	&12.365	&9.842 &12.503 &9.866\\
\hline
\end{tabular}
\end{center}

\vspace{-5pt}

\caption{Mean/patch-wise/block-wise/pixel-wise AUSE and correlation between $L_1$ loss and uncertainty, and NLL on NYU Depth Dataset V2. Our infer-transformation, infer-dropout and infer-noise are compared with MC-dropout~\citep{gal2016dropout}. MC-drop$^1$ uses the output of the original model as prediction while MC-drop$^2$ uses the mean of output samples from the re-trained model (with added dropout) as prediction. Models evaluated: FCRN model.}

\vspace{-10pt}
\label{Tab: depth estimation: correlation with error}
\end{table*}
\subsection{Experiment Results}\label{sec:res}

\textbf{Qualitative Results.} \figref{fig:variance_map} shows some qualitative results for an example image in the SR task. We can see that the variance maps generated in our task are consistent with the level of ambiguity.  Specifically, in our methods, high variance occurs in areas with high randomness and high frequency. For the depth estimation task shown in \figref{fig:variance_map_depth}, high variance usually occurs in the area with high spatial resolution and large depth. \textcolor{black}{As expected, these high-variance areas usually correspond to large prediction errors.}

\textbf{Evaluation Metrics.} 
Commonly used metrics to evaluate uncertainty estimation include Brier score (BS), expected calibration error (ECE), and negative log-likelihood (NLL)~\citep{deepensemble,TS}. However, BS and ECE are for classification tasks only and hence not applicable in our setting. We therefore use the following metrics for evaluations: (1) \textbf{NLL}, which is defined in regression tasks by assuming a Gaussian distribution. However, note that NLL depends on not only the quality of uncertainty estimation but also the prediction accuracy itself. Therefore contrary to previous belief, we argue that it is not an ideal metric for evaluating uncertainty estimation. (2) \textbf{Area Under the Sparsification Error (AUSE)}, \textcolor{black}{which quantifies how much uncertainty estimation coincides with the true errors \citep{AUSE}. (3) \textbf{Correlation} between the estimated uncertainty and the error. Here we define four variants of correlation (\textcolor{black}{see details in the Supplement)}: \emph{pixel-wise}, \emph{mean}, \emph{block-wise}, and \emph{patch-wise} correlations to evaluate performance at the pixel, image, block, and patch levels, respectively. 
The intuition is that in many situations it is more instructive and meaningful when uncertainty is visualized in each region (e.g. a region with a possible tumor for a medical imaging application). 
Note that block-wise correlation depends on specific segmentation algorithms, while patch-wise correlation defines regions in an algorithm-independent way. Similarly, we also define four evaluation forms for AUSE.}

\textcolor{black}{For our training-free methods, these metrics are computed between uncertainty and the error from the \emph{original} model (without perturbation), because we will still use the original model for prediction. For training-required methods such as MC-dropout (i.e. MC-drop$^2$ in Table~\ref{Tab:correlation with error})~\citep{gal2016dropout}, deep ensemble~\citep{deepensemble} and \textcolor{black}{log likelihood maximization} (LLM)~\citep{zhang2019reducing, poggi2020uncertainty}, the mean of output samples are used as prediction. 
Meanwhile, we also evaluate another MC-dropout variant, denoted as MC-drop$^1$, where the output of the original model is used as prediction, to be consistent with training-free methods. The evaluation results using metrics described above are shown in Table~\ref{Tab:correlation with error} and Table~\ref{Tab: depth estimation: correlation with error}. }

\begin{figure*}[t]
\centering
	 \includegraphics[width=0.9\linewidth]{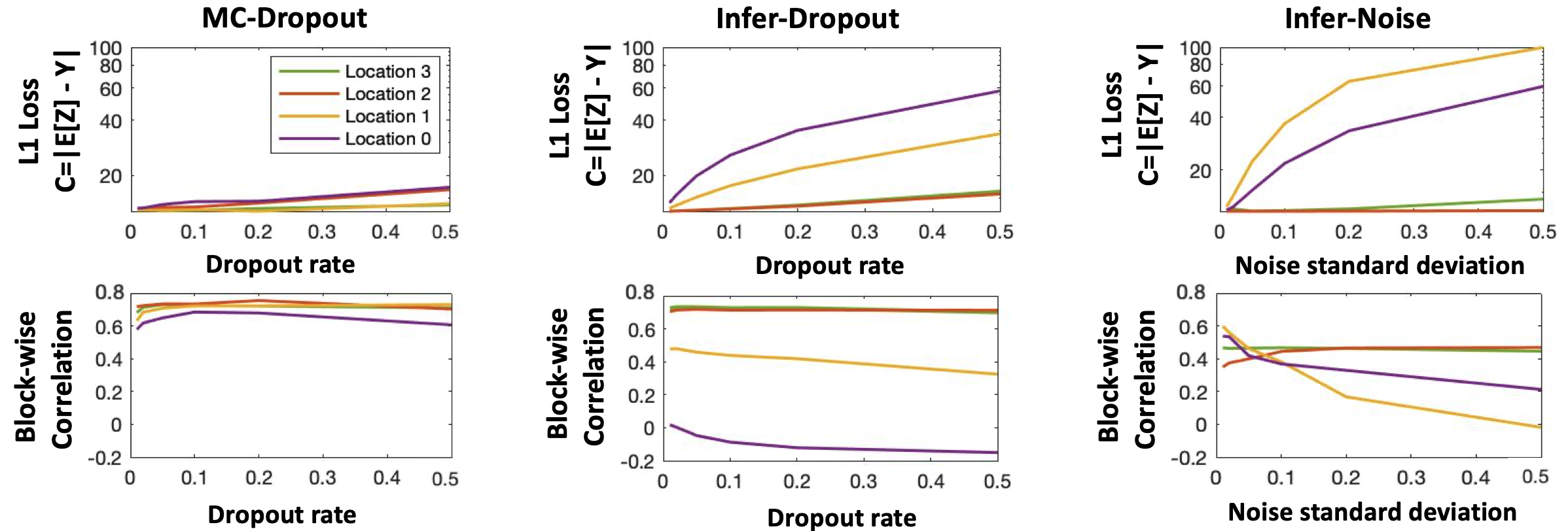}
	 \caption{\textcolor{black}{\textbf{Top}: $L_1$ loss of perturbed model for MC-dropout~\citep{gal2016dropout}, infer-dropout and infer-noise. Various dropout rates and noise levels have been evaluated. Location 0 is right after the input; location 1, 2, 3 are intermediate layers. \textbf{Bottom}: Correlation between error and variance with different locations and perturbation strength. For infer-dropout, note that location 2 and 3 cause minimal increase in the predictive performance $C =|\mathbb{E}[Z] - Y|$ measured by $L_1$ loss after perturbation (i.e. high tolerability), leading to high correlation.}}
\label{fig:correlation_error}
\end{figure*}

\textbf{The Role of Tolerable Perturbations.}
Tolerable perturbations play a crucial role in obtaining effective uncertainty estimation. Better tolerability means smaller decrease in accuracy after perturbation.
\figref{fig:correlation_error} shows the correlation for different amount of perturbations (noise or dropout) in different locations, \textcolor{black}{and the corresponding predictive performance $C = |\mathbb{E}[Z] - Y|$ (evaluated as $L_1$ loss) \emph{after perturbations}}. As we can see, the optimal cases to generate uncertainty maps with high correlation require that \textcolor{black}{$C$} should remain small after perturbation (high tolerability). Generally, our experiments suggest that perturbations leading to less than $20\%$ relative drop of performance work well for uncertainty estimation. This observation verifies \thmref{thm:corr}. Specifically, note that by definition (in the `Notations' paragraph), $\mathbb{E}[Z]=\mu(X)$; therefore given the `random bias' assumption (in the `Assumptions' paragraph) that $Y - \mu(X) \sim \N(0,B^2)$, the expectation of $C$ over all the pixels $E[C]=\mathbb{E}[|\mathbb{E}[Z] - Y|]=\mathbb{E}_X[|\mu(X) - Y|]=\sqrt{2/\pi}B$. By Remark 2 of \thmref{thm:corr}, $\rho(\sigma,e)$ is monotonically decreasing w.r.t. to $B$, which is linear to $E[C]$. 

Interestingly, different methods have different ways of achieving high tolerability: (1) For MC-dropout, involving dropout during training increases the robustness of models against perturbations, keeping the loss relatively small after adding dropout layer in most locations during inference; (2) for infer-dropout, adding dropout layer in intermediate locations (i.e. location 2 and location 3) where the information is the most redundant~\citep{he2014reshaping}, can effectively alleviate disturbance; (3) for infer-noise, adding noise with small standard deviation effectively limits the perturbation level. More interestingly, we further find that for both MC-dropout and infer-dropout, adding perturbation in intermediate layers is usually the optimal choice for uncertainty estimation. Applying infer-dropout in these intermediate layers, we could achieve comparable or even better performance compared to training-required \textcolor{black}{baselines}. For infer-noise, locations do not have a similar effect; one can therefore further tune the noise strength $\sigma$ to achieve a higher correlation.

\section{Applications} 
\label{sec: app}

We find several applications that can be benefited from the uncertainty estimated in our methods. The first application is to improve the quality of SR results. We propose a novel method that takes the pixel-wise uncertainty map as a weight term for the regression loss while keeping the original adversarial loss; this could provide a more photo-realistic SR output with finer structures and sharper edges. Another application is active learning~\citep{gal2017deep}, which aims to use uncertainty to choose the next batch of data for annotation.
Our result shows that active learning based on our generated uncertainty maps can provide a higher data efficiency (details are shown in the Supplement).

\vspace{-5pt}
\section{Conclusion \& Discussion}
\label{sec: conclusion}
 In this work, \textcolor{black}{we perform a systematic exploration into training-free uncertainty estimation for dense regression, an unrecognized yet important problem, and provide a theoretical construction justifying such estimations}. We propose three simple, scalable, and effective methods, i.e. \textit{infer-transformation}, \textit{infer-noise}, and \textit{infer-dropout}, for uncertainty estimation in both black-box and gray-box cases. Surprisingly, our training-free methods achieve comparable or even better results compared to training-required state-of-the-art methods. Furthermore, we demonstrate adding tolerable perturbations is the key to generating high-quality uncertainty maps for all methods we studied. Meanwhile, we find the optimal choice of the training-free uncertainty estimation method is task-dependent. We suggest using infer-transformation for super-resolution and infer-noise for depth estimation. The limitation of our works is the requirement of well-trained neural networks. And our methods are not applicable to deep neural networks with random guesses.  
 
\section{Acknowledgement}
The authors thank the reviewers/SPC/AC for the constructive comments to improve the paper. We thank Intel, Analog Devices, Thermo Fisher Scientific, NSF grants 1563880 and 1607189 to support this project. Lu Mi is partially supported by MathWorks Fellowship. Hao Wang is partially supported by NSF Grant IIS-2127918 and an Amazon Faculty Research Award.

\bibliography{reference}

\section{Appendix}
\section{Proof of Theorem 1}
\label{sec: proof of theorem 3.1}

\begin{proof}
\begin{align*}
\E [e] &= \E_{\sigma} \E[e | \sigma] 
= \E_{\sigma} \E\left[|\sigma \eone - B\etwo| | \sigma \right]
\\&= \E_{\sigma} \E\left[|\N(0, \sigma^2+B^2)| | \sigma \right]
\\&= \E_{\sigma} \sqrt{\frac{2}{\pi}} \sqrt{\sigma^2+B^2} = \sqrt{\frac{2}{\pi}} \E[\sigma_B]
\end{align*}
\begin{align*}
\E [\sigma e] &
= \E_{\sigma} \E\left[|\sigma^2 \eone - \sigma B\etwo| | \sigma \right]
\\&= \E_{\sigma} \E\left[|\N(0, \sigma^2(\sigma^2+B^2))| | \sigma \right]
\\&= \E_{\sigma} [\sqrt{\frac{2}{\pi}} \sigma \sqrt{\sigma^2+B^2} ]= \sqrt{\frac{2}{\pi}} \E[\sigma \sigma_B]
\end{align*}
\begin{align*}
\E [e^2] &
= \E_{\sigma} \E\left[(\sigma \eone - B\etwo)^2 | \sigma \right]
= \E_{\sigma} [\sigma^2+B^2 ]
= \E [\sigma_B^2]
\end{align*}
Putting the equations above together, we have
\begin{align*}
\rho(\sigma,e) &= \frac{\E[\sigma e] - \E[\sigma] \E [e]}{\sqrt{\E[\sigma^2]-(\E[\sigma])^2}\sqrt{\E[e^2]-(\E[e])^2}}
\\&= \frac{\sqrt{2/\pi}(\E[\sigma \sigma_B] - \E[\sigma] \E [\sigma_B]) }{\sqrt{\V[\sigma]}\sqrt{\E[\sigma_B^2]-\frac{2}{\pi}(\E[\sigma_B])^2}}
\\&= \sqrt{\frac{2}{\pi}}\sqrt{\frac{\E[\sigma_B^2]-(\E[\sigma_B])^2}{\E[\sigma_B^2]-\frac{2}{\pi}(\E[\sigma_B])^2}}\frac{\E[\sigma \sigma_B] - \E[\sigma] \E [\sigma_B]}{\sqrt{\V[\sigma]}\sqrt{\V[\sigma_B]}}
\\&= \sqrt{\frac{2}{\pi} \cdot \frac{1 - \lambda^2}{1 - \frac{2}{\pi}\lambda^2}} \rho(\sigma, \sigma_B)
\end{align*}
\end{proof}

\section{Assumption Evaluation for Correlation Analysis}

\begin{figure*}[t]
\centering
	 \includegraphics[width=0.75\linewidth]{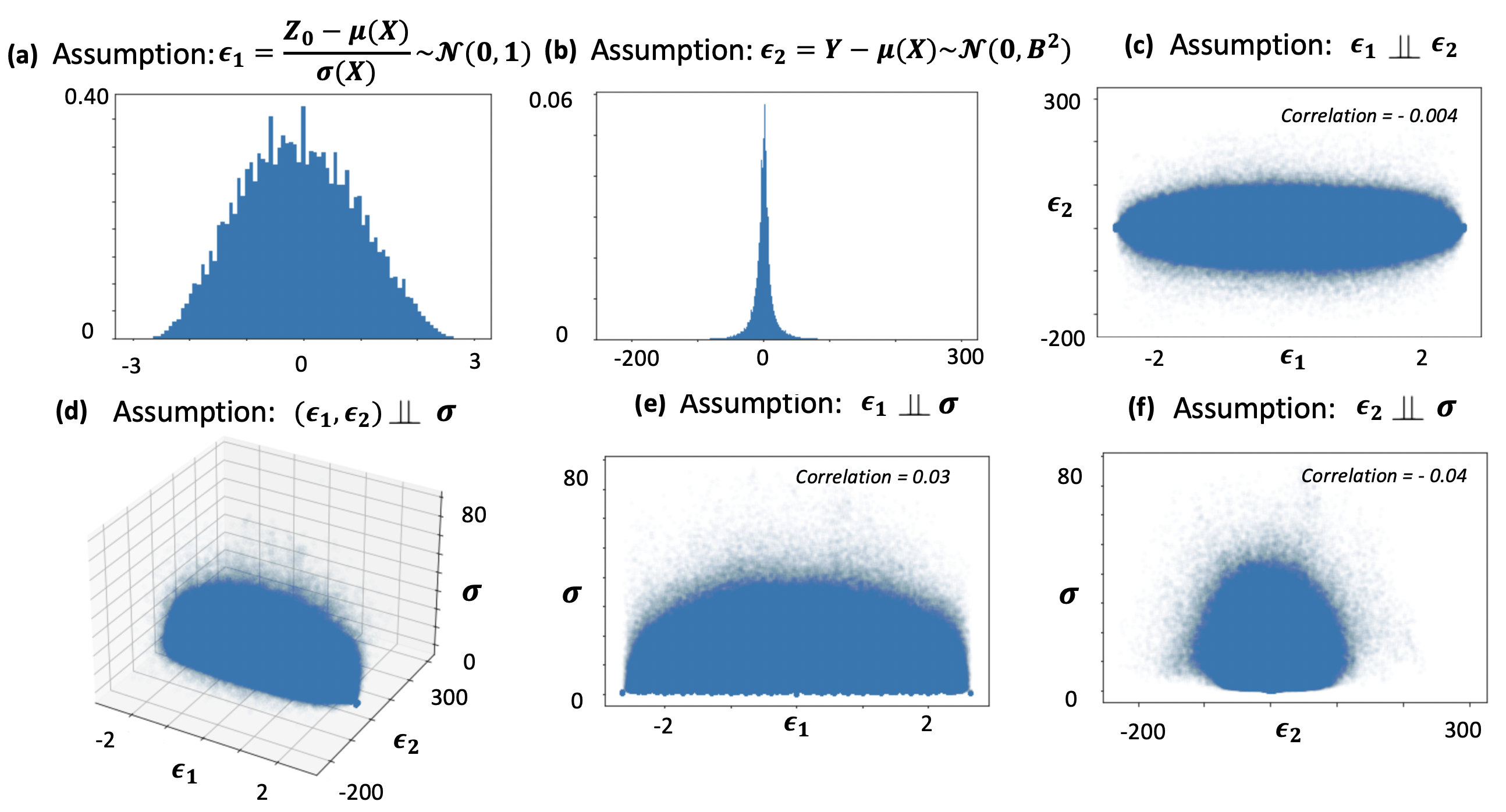}
	 \caption{Evaluated assumptions for correlation analysis including (a) a histogram plot of $\epsilon_1$ to demonstrate heterogeneous perturbation, (b) a histogram plot of $\epsilon_2$ to demonstrate random bias, and (c) a scatter plot of $\epsilon_1$ and $\epsilon_2$ to demonstrate the independence, and (d), (e), (f) scatter plots between $\epsilon_1$, $\sigma$ and $\epsilon_2$ to demonstrate the independence. Values  of correlation are also illustrated.}

\label{fig:assumption_evaluation}
\end{figure*}

In this section, we evaluate the assumptions to perform our correlation analysis, and find most of them are satisfied to some degree. We demonstrate an example of described variables in the assumption for infer-transformation on SRGAN model in Figure~\ref{fig:assumption_evaluation}. In (a), we show the histogram of $\epsilon_1 = \frac{Z_0 - \mu(X)}{\sigma(X)}$ of all pixels in the Set14, and such distribution is very close to a standard normal distribution $\mathcal{N} (0,1)$ as expected, which demonstrates our Assumption 1 (in the main paper) related to heterogeneous perturbation is satisfied. Similarly, we evaluate the random bias assumption (Assumption 2) by showing the histogram of $\epsilon_2 = Y - \mu(X)$ in (b). We find this distribution is relatively close to a normal distribution $\mathcal{N} (0,B^2)$, and here the measured $B$ is $16.1$. In the meanwhile, we also evaluate the dependence of these random variables (Assumption 3). Figure~\ref{fig:assumption_evaluation}(c) shows the independence assumption of $\epsilon_1$ and $\epsilon_2$ is satisfied to some degree, and the independence assumption of $(\epsilon_1,\epsilon_2)$ and $\sigma$ is evaluated in (d), (e), (f).

\section{Applications Benefit from Uncertainty Estimation}
\label{sec: applications}

\begin{figure*}[t]
\centering
	 \includegraphics[width=0.9\linewidth]{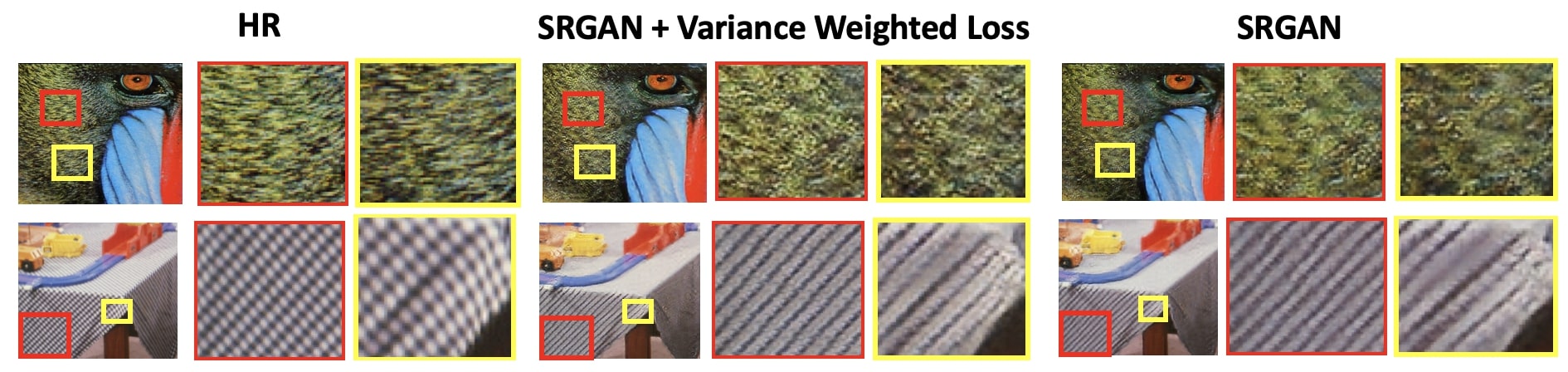}
	 \vskip -0.2cm
	 \caption{The first application using uncertainty map estimated in our methods to improve the quality of SR results. We compare SR results that use $L_2$ loss re-weighted by variance map \textbf{(middle)} and that do not \textbf{(right)}. HR \textbf{(left)} represents high resolution image. Results evaluated in Set 14.}
\label{fig:application}
\end{figure*}

The first application is to improve the quality of SR results. We propose a novel and efficient method which takes the pixel-wise uncertainty map as a weight term for the regression loss, while keeps the original adversarial loss, which could provide a more photo-realistic SR output with finer structures and sharper edges, shown in Fig.~\ref{fig:application}. Another application is active learning~\citep{gal2017deep}, which aims to use uncertainty to guide annotations, then only a small subset of data are required to improve training. We find active learning based on our generated uncertainty maps can improve the performance with more efficiency, shown in Table~\ref{Tab:active learning}.

\begin{table}[ht]
\begin{center}
\begin{tabular}{c|ccc}
\hline
Method &SSIM &PSNR &$L_1$ \\
\hline
original &0.772 &23.841 &11.623\\
random &0.773 &23.616 &12.344 \\
high uncertainty &\textbf{0.790} &\textbf{23.942} &\textbf{11.497}\\

\hline

\end{tabular}
\end{center}

\caption{Using active learning on our generated uncertainty maps can provide a higher data efficiency. Here we select samples with high uncertainty yields better results than select randomly. }
\vspace{-10pt}
\label{Tab:active learning}
\end{table}

\section{Details on Network Noise/Dropout Injection}
\label{sec: network location}

To perform uncertainty estimation using infer-noise, infer-dropout, and baseline MC-dropout on both SRGAN model and SRresnet model, we choose 4 different locations for noise injection, including the layers right after the input, as well as some intermediate layers, as shown in \figref{appendix:network_structure}.

\begin{figure*}[t]
\centering
\includegraphics[width=0.65\linewidth]{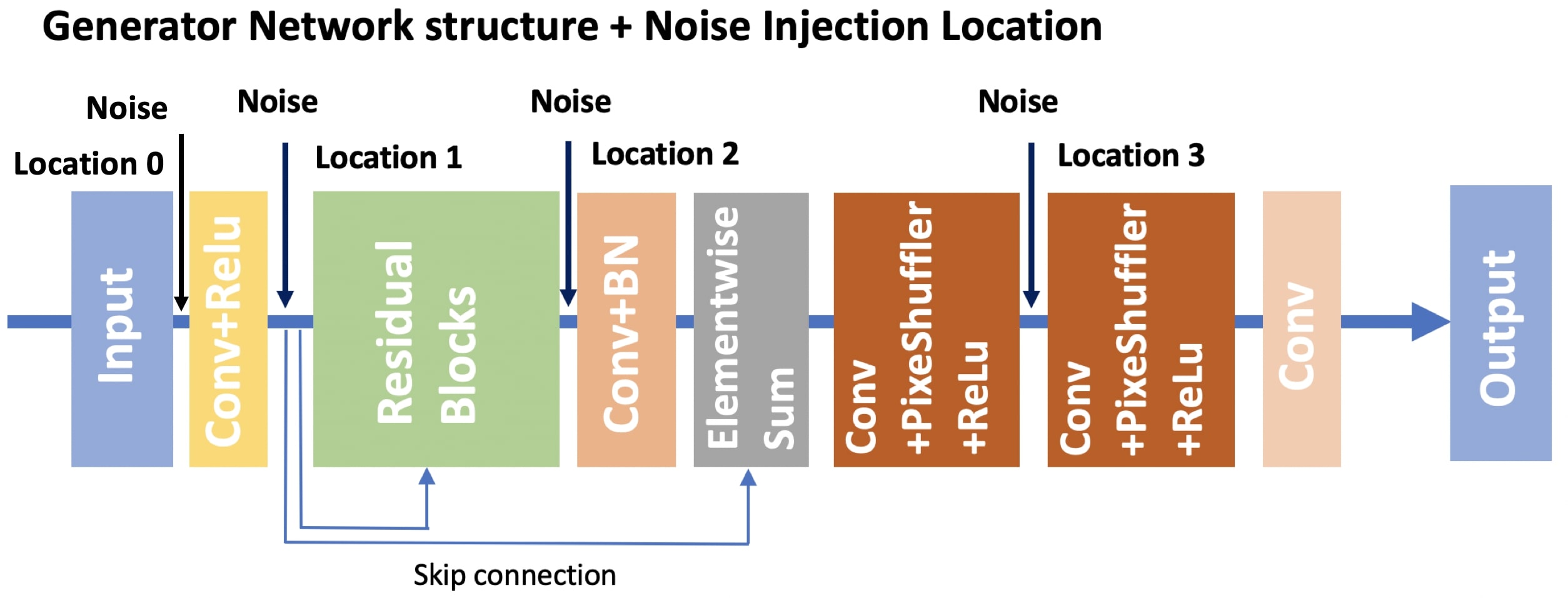}
\caption{Different locations for infer-noise and infer-dropout in SRGAN and SRresnet for super resolution. For each experiment, the noise or dropout is injected at a single location with one perturbation level.}
\label{appendix:network_structure}
\end{figure*}

\begin{figure*}[h!]
\centering
	 \includegraphics[width=0.65\linewidth]{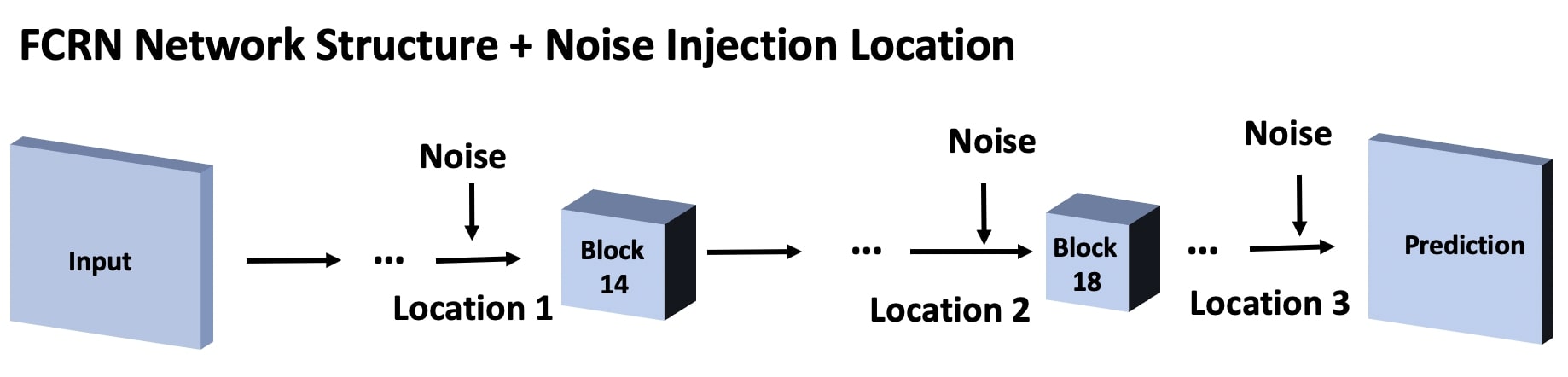}
	 \caption{Different locations for noise or dropout injection in the FCRN model for depth estimation.}

\label{fig:FCRN_network}
\end{figure*}

To perform uncertainty estimation using infer-noise and infer-dropout on FCRN model, we choose 3 different locations for noise injection, including the layers right before the output, as well as some intermediate layers, as shown in \figref{fig:FCRN_network}. For baseline MC-dropout, we choose location 3 where the dropout layer added during training for the original model.

\section{Tolerability on SRresnet and FCRN}
\label{sec: Tolerability on SRresnet and FCRN}

\begin{figure*}[t]
\centering
	 \includegraphics[width=0.85\linewidth]{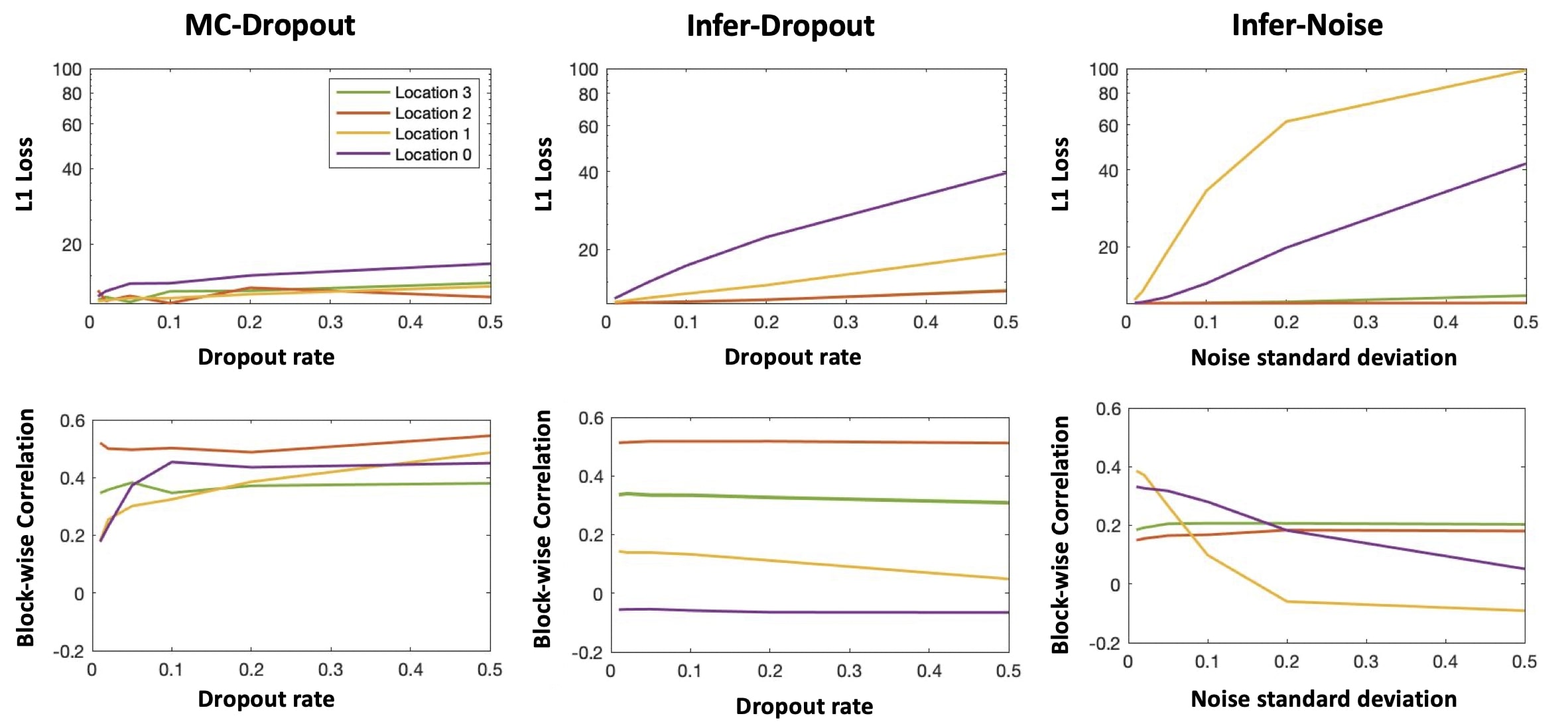}
	 \caption{Evaluation on SRresnet model for super resolution. \textbf{Top}: The predictive performance $C =|\mathbb{E}[Z] - Y|$ of the perturbed model represented by $L_1$ loss for infer-dropout and infer-noise and baseline MC-dropout~\cite{gal2016dropout}. Various dropout rates and noise levels have been evaluated. Location 0 is right after the input; location 1, 2, 3 are intermediate layers. \textbf{Bottom}: Correlation between error and variance with different locations and perturbation strength. Note that location 2 cause minimal increase in the predictive performance $C =|\mathbb{E}[Z] - Y|$ represented by $L_1$ loss after perturbation (i.e., high tolerability), leading to high correlation for MC-dropout and infer-dropout.}
\label{fig:correlation_srresnet_error_figure}
\end{figure*}

\begin{figure*}[h!]
\centering
	 \includegraphics[width=0.85\linewidth]{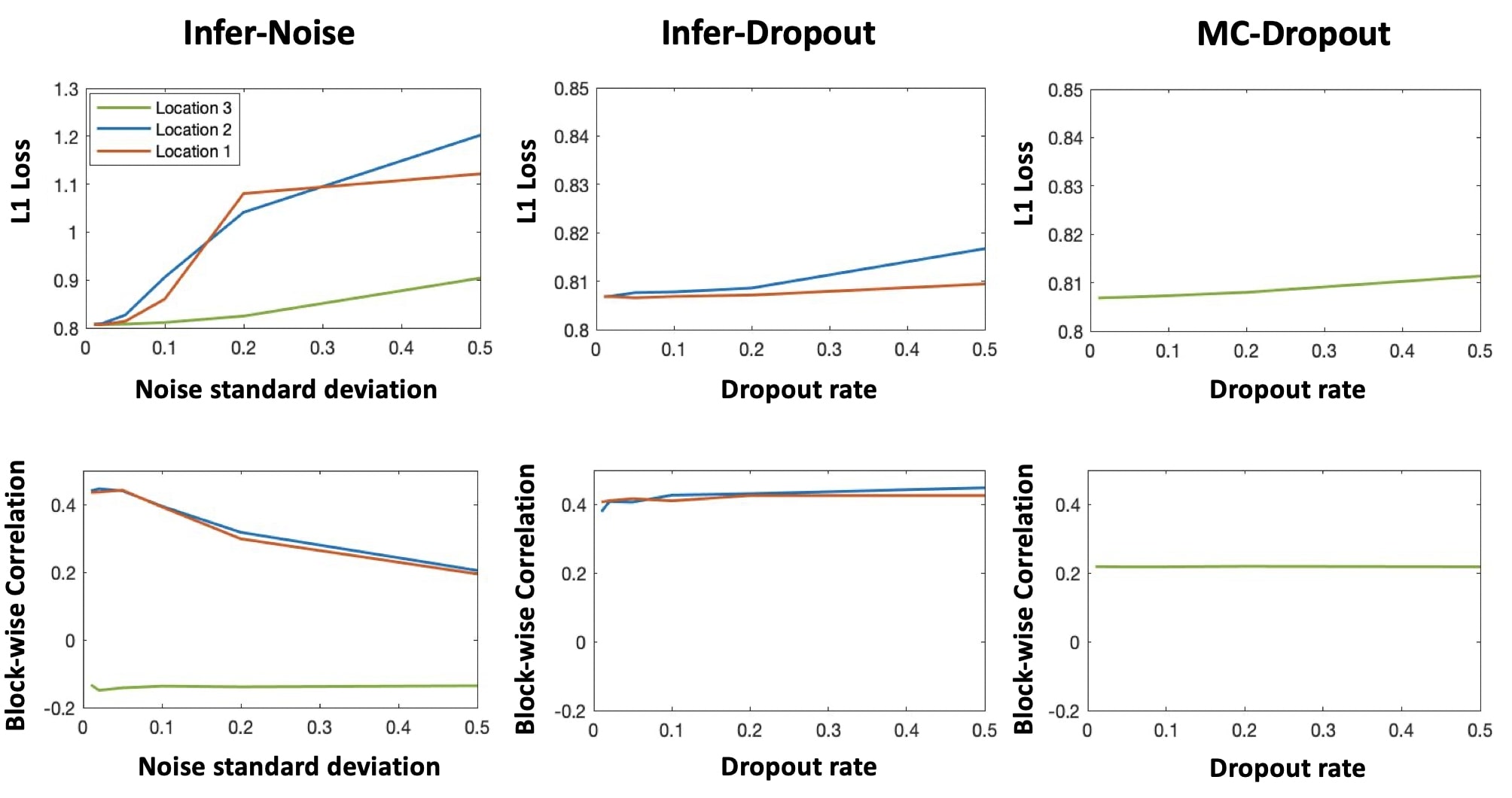}
	 \caption{Evaluation on FCRN model for depth estimation. \textbf{Top}: The predictive performance $C =|\mathbb{E}[Z] - Y|$ of the perturbed model represented by $L_1$ loss for infer-dropout and infer-noise and baseline MC-dropout~\cite{gal2016dropout}. Various dropout rates and noise levels have been evaluated. Location 1, 2 are intermediate layers, location 3 is right before the last convolutional layer. \textbf{Bottom}: Correlation between error and variance with different locations and perturbation strength. Note that location 1 and 2 cause minimal increase in the predictive performance $C =|\mathbb{E}[Z] - Y|$ represented by $L_1$ loss after perturbation (i.e., high tolerability), leading to high correlation for infer-dropout.}
\vspace{-5pt}
\label{appendix:correlation_depth_figure}
\end{figure*}

\figref{fig:correlation_srresnet_error_figure} shows the perturbation on the SRresnet model, and \figref{appendix:correlation_depth_figure} shows the  perturbation on the FCRN model for depth estimation. We plot the correlation between different amount of perturbations (noise or dropout) in different locations \textcolor{black}{and the corresponding predictive performance (evaluated as $L_1$ loss) \emph{after perturbations}}. For SRresnet model, We find that for both MC-dropout and infer-dropout, adding perturbation in intermediate layers is usually the optimal choice for uncertainty estimation. For infer-noise, locations do not have similar effect; one can therefore further tune the noise strength $\sigma$ to achieve higher correlation. For FCRN model, when infer-dropout or infer-noise is applied, intermediate layers are also usually the optimal choices for uncertainty estimation. The results are consistent with that of SRGAN model.

\begin{figure*}[t]
\centering
\includegraphics[width=0.6\linewidth]{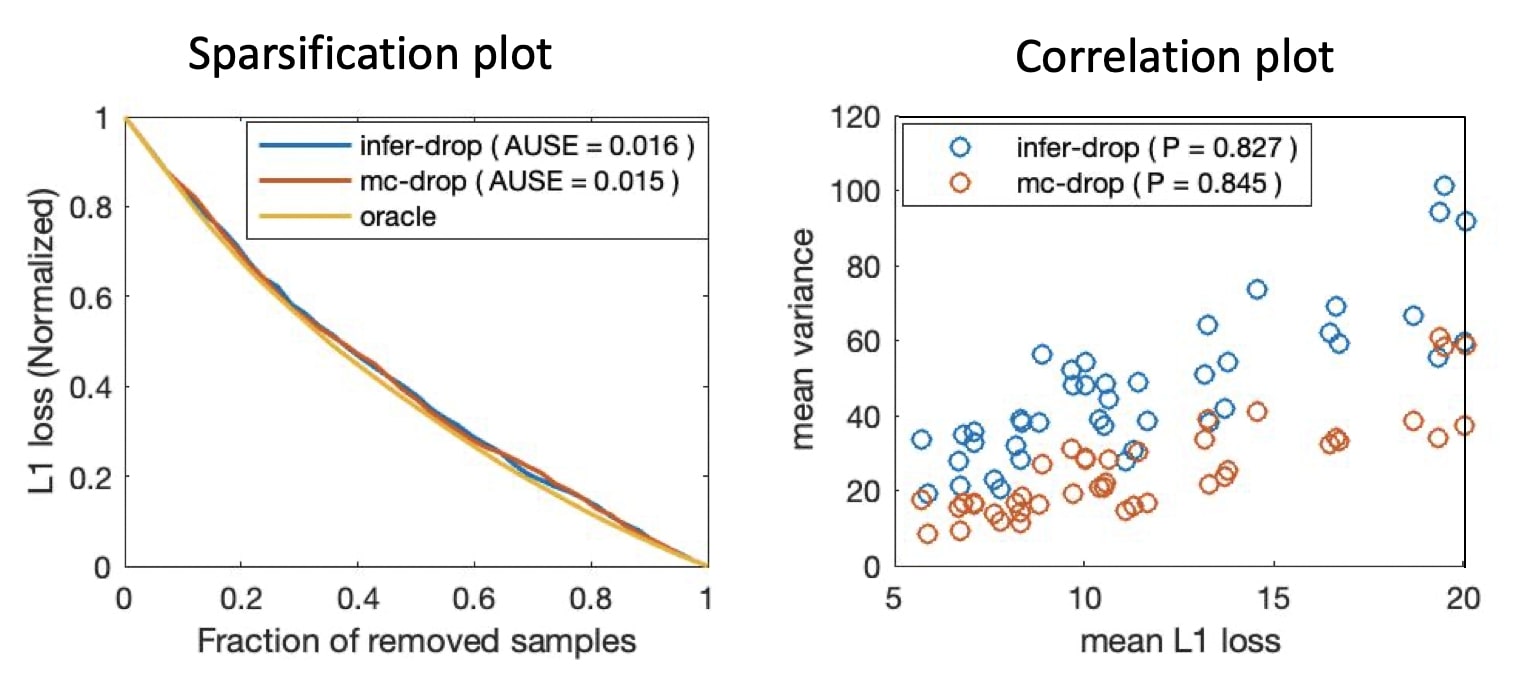}
\caption{\textbf{Left:} The sparsification error plot using the mean of uncertainty and $L_1$ loss of each image sample for infer-dropout and MC-dropout, the values of AUSE are also presented. \textbf{Right:} The scatter plot using the mean of uncertainty and $L_1$ loss of each image sample, the values of mean $L_1$ correlation are also presented.}
\label{fig: metrics}
\end{figure*}

\section{Details on Evaluation Metrics}
\label{sec: metrics}
In this section, we provide the details for our proposed correlation-based evaluation metrics.
\label{appendix: Evaluation Metrics}
Assuming we have N outputs given the same input $x$ from our infer-transformation, infer-drop and infer-noise, each output is represented by $Y_w$. Given the output image with the size of $H\times W$ , the error we define for regression task is pixel-wise $L_1$ loss and $L_2$ loss, represented by $L_{1,ij}$ and $L_{2,ij}$, where $i$, $j$ is the corresponding coordinates of the pixel $P_{ij}$ in the output image. The uncertainty (variance) estimated in these methods is also a pixel-wise value, represented by $V_{ij} = \frac{\sum_{w=1}^{N}(Y_{w,ij} - \overline{Y}_{ij})^2} {N}$.
The pixel-wise $L_1$ correlation is defined as $corr(\{V_{ij}\}, \{L_{1,ij}\})$. The second metric is mean correlation, the mean $L_1$ error $\overline{L}_{1,z} = \frac{\sum_{i=1}^{W}\sum_{j=1}^{H} L_{1,ij}}{W\times H}$ is defined as the average error of a single image $z$, correspondingly, the mean variance is defined as $\overline{V}_{z} = \frac{\sum_{i=1}^{W}\sum_{j=1}^{H} V_{ij}}{W\times H}$, the mean $L_1$ correlation is defined as $corr(\{\overline{V}_{z}\}, \{\overline{L}_{1,z}\})$. This metric has been used in~\cite{zhang2019reducing}. The third metric for evaluation is the block-wise correlation -- a new metric we propose in this work. To compute block-wise correlation, we need to firstly apply a local segmentation algorithm to the output of the trained model to cluster pixels with similar low-level context. Here we use the local-center-of-mass approach~\cite{aganj2018unsupervised} \textcolor{black}{to perform segmentation}. We denote each cluster as $C_i$. The variance of $K_{C_i}$ pixels inside each cluster (block) $C_i$ is then averaged and replaced with the mean value $\widetilde{V}_{i} = \frac{\sum_{P_{ij} \in C_i}^{K_{C_i}} V_{ij}}{K_{C_i}}$. The block-wise $L_1$ loss $\widetilde{L}_{1,i}$ can be calculated similarly. After that, we calculate the pixel-wise correlation of each pixels with the updated value as the $L_1$ block-wise correlation $corr(\{\widetilde{V}_{i}\},\{\widetilde{L}_{1,i}\})$. For the fourth metric, patch-wise correlation, where the segmentation clusters in block-wise correlation are replaced by patches. In our analysis, each image is divided into $10 \times 10$ patches. And then the patch-wise correlation is calculated with following the same rule as block-wise correlation. Besides correlation, we also define four similar metrics in terms of AUSE. More details related with the definition of AUSE are in~\cite{AUSE}.

Meanwhile, as illustrated in Fig.~\ref{fig: metrics}, we find that sparsification error has a strong association with correlation, when the oracle sparsifications of different methods are the same. As a result, when AUSE of infer-dropout and MC-dropout (defined as MC-drop$^1$ here) is nearly identical, correlation is almost the same.

\begin{table*}[t]
\begin{center}
\begin{tabular}{c|cc|cc|cc}
\hline
&\multicolumn{2}{c|}{Densenet} &\multicolumn{2}{c|}{UNET} &\multicolumn{2}{c}{Resnet}\\
\hline
&Entropy &MC-drop &Entropy &MC-drop &Entropy &MC-drop\\
\hline
Mean Correlation&\textbf{0.928} &0.718 &\textbf{0.964} &0.881 &\textbf{0.669} &0.537\\
Pixel-wise Correlation &\textbf{0.502} &0.209 &\textbf{0.789} &0.317 &-- &--\\
\hline
\end{tabular}
\end{center}
\caption{Correlation of uncertainty and cross-entropy loss, comparing using entropy with the baseline MC-dropout, models evaluated are Densenet for the segmentation on CamVid dataset, UNET for segmentation on SNEMI3D dataset and Resnet for classification on CIFAR100 dataset.}
\label{Tab: classification: correlation with cross-entropy loss}
\end{table*}

\section{Evaluations on Classification Tasks}
\label{appendix: Classification Tasks}
We compare the simple training-free method using entropy and the training-required method MC-dropout on classification tasks. For classification tasks, the most straightforward and commonly used method is to calculate the entropy of output probability as uncertainty, which already qualifies as a training-free method. We then compare it with a sampling-based and training-required method -- MC-dropout, tuned on different locations and using 8 samples. Here we conduct three experiments: the first one is multi-class segmentation task using Densenet~\cite{huang2017densely} on CamVid dataset; the second one is a binary segmentation task using UNET~\cite{ronneberger2015u} on a biomedical public benchmark dataset from the SNEMI3D challenge; and the third one is a classification task on CIFAR100 using ResNet~\cite{he2016deep}. We calculate the correlation between the entropy of softmax output and the cross-entropy loss. We find that using entropy outperforms MC-dropout based on the correlation metric, as shown in Table~\ref{Tab: classification: correlation with cross-entropy loss}.

\section{Visualization of Uncertainty Maps}
\label{appendix: more experiment results}
\begin{figure*}[t]
\centering
	 \includegraphics[width=1\linewidth]{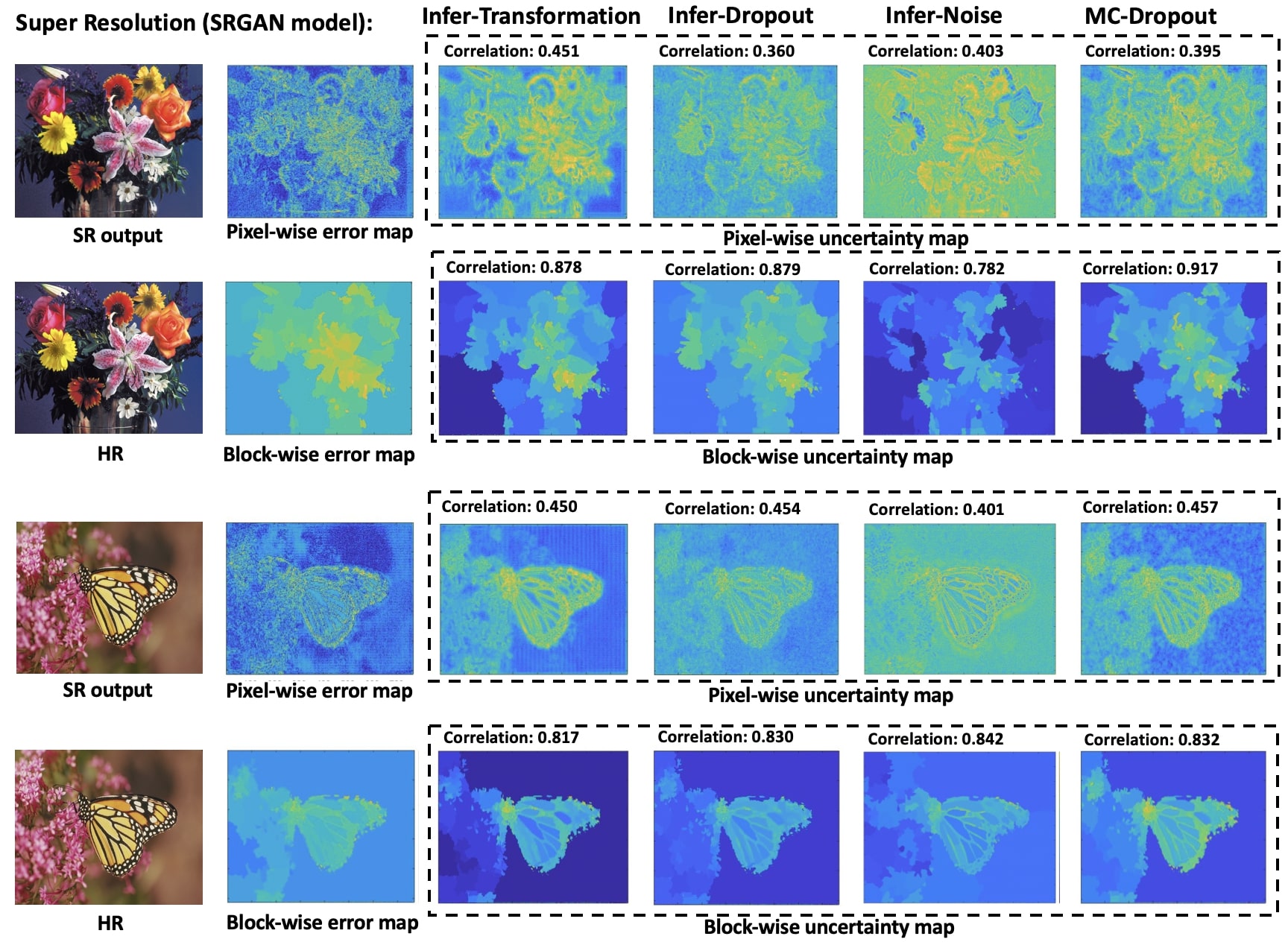}
	 \caption{Visualization of uncertainty maps (log scale) and error map (log scale) from infer-transformation, infer-dropout, infer-noise compared with the baseline MC-dropout, evaluated on the SRGAN model on Set14 dataset for super-resolution task.}
\label{fig:SRGAN_visualization}
\end{figure*}

\begin{figure*}[t]
\centering
\includegraphics[width=1\linewidth]{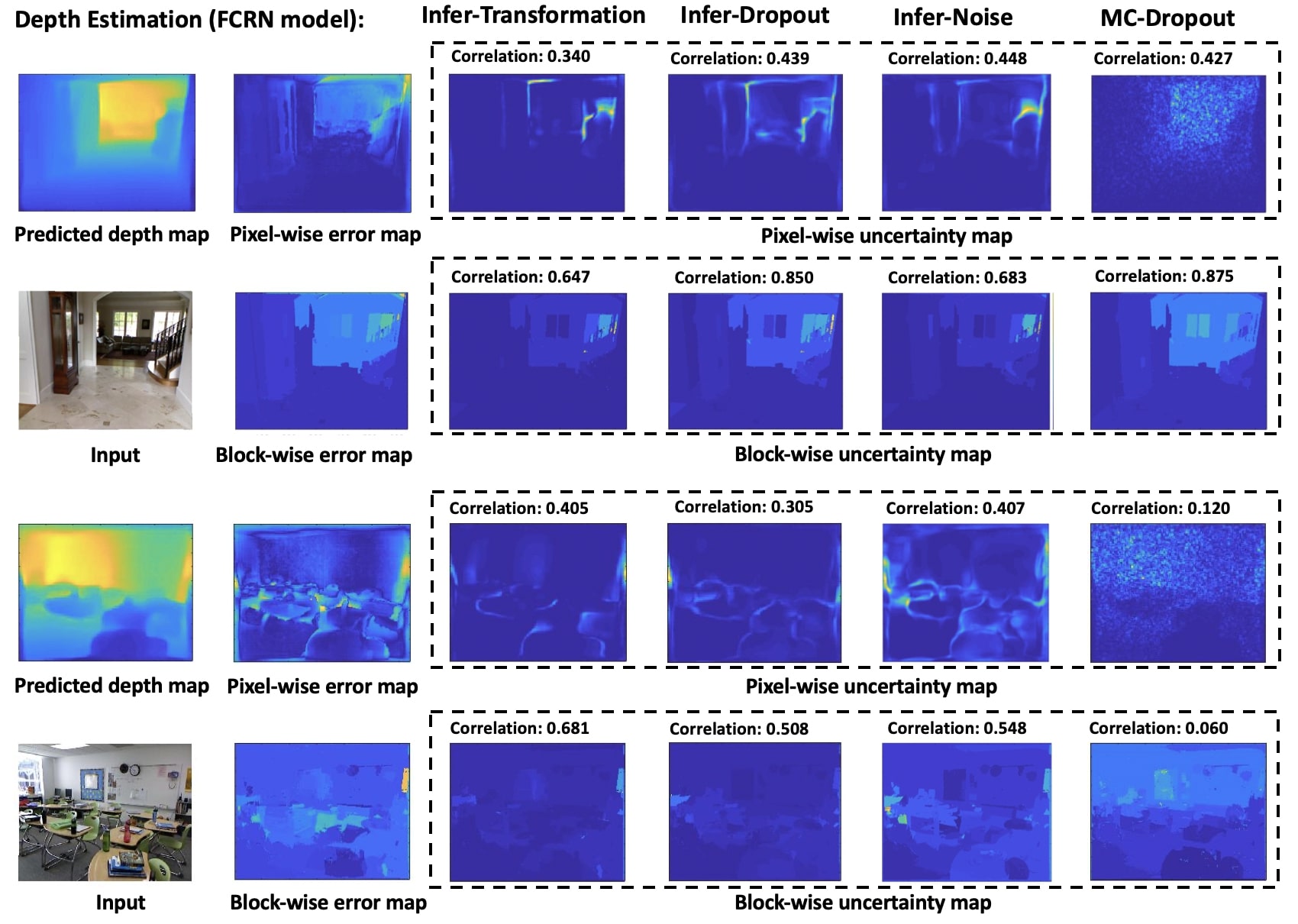}
	 \caption{Visualization of uncertainty maps and the error map from infer-transformation, infer-dropout, infer-noise compared with the baseline MC-dropout, evaluated on the FCRN model on NYU depth dataset V2 for the depth estimation task.}
\label{fig:depth_visualization}
\end{figure*}

The uncertainty maps generated using infer-transformation, infer-dropout, infer-noise compared with MC-dropout, and the corresponding error maps for images are shown in Fig.~\ref{fig:SRGAN_visualization} for the super-resolution task, and in Fig.~\ref{fig:depth_visualization} for the depth estimation task. 

One interesting observation is that MC-dropout tends to capture local variance and ignore high-level semantics, partially because the dropout layer is always at the end of the network. As a result, it is difficult for MC-dropout to produce uncertainty estimates in detail-rich regions. For example, the input image in the bottom row of Fig.~\ref{fig:depth_visualization} contains a lot of details with chairs and desks. Unfortunately MC-dropout tends to ignore these details and only produce high variance in the upper half of the image (region with large depth), leading to poor correlation.

\label{appendix: Visualization of depth Uncertainty Map}

\end{document}